\documentclass[11pt]{article}

\usepackage[preprint]{acl}

\usepackage{times}
\usepackage{latexsym}

\usepackage[T1]{fontenc}
\usepackage[utf8]{inputenc}

\usepackage{microtype}

\usepackage{inconsolata}

\usepackage{graphicx}
\usepackage{hyperref}       % hyperlinks
\usepackage{subcaption}    % for subfigure environment
\usepackage{url}            % simple URL typesetting
\usepackage{booktabs}       % professional-quality tables
\usepackage{amsfonts}       % blackboard math symbols
\usepackage{nicefrac}       % compact symbols for 1/2, etc.
\usepackage{graphicx}
\usepackage{microtype}      % microtypography
\usepackage{xcolor}
\usepackage{colortbl}
\definecolor{lightred}{RGB}{255, 204, 204}
\definecolor{lightgreen}{RGB}{204, 255, 204}
\usepackage{float}
\usepackage{enumitem} 

\usepackage{amsmath}
\usepackage{adjustbox}
\usepackage{multirow}
\usepackage{tabularx}
\usepackage{array}

\title{How Do Inpainting Artifacts Propagate to Language?}

\author{
 \textbf{Pratham Yashwante\textsuperscript{1}},
 \textbf{Davit Abrahamyan*\textsuperscript{1}},
 \textbf{Shresth Grover*\textsuperscript{1}},
 \textbf{Sukruth Rao*\textsuperscript{1}}
\\
\\
 \textsuperscript{1}UC San Diego
\\
 \small{
   \textbf{Correspondence:} \href{mailto:pyashwante@ucsd.edu}{pyashwante@ucsd.edu}
 }
}

\begin{document}
\maketitle
\begin{abstract}
We study how visual artifacts introduced by diffusion-based inpainting affect language generation in vision-language models.
We use a two-stage diagnostic setup in which masked image regions are reconstructed and then provided to captioning models, enabling controlled comparisons between captions generated from original and reconstructed inputs.
Across multiple datasets, we analyze the relationship between reconstruction fidelity and downstream caption quality.
We observe consistent associations between pixel-level and perceptual reconstruction metrics and both lexical and semantic captioning performance.
Additional analysis of intermediate visual representations and attention patterns shows that inpainting artifacts lead to systematic, layer-dependent changes in model behavior.
Together, these results provide a practical diagnostic framework for examining how visual reconstruction quality influences language generation in multimodal systems.
\end{abstract}

\section{Introduction}

\begin{figure*}[ht]
    \centering
    \setlength{\tabcolsep}{6pt}

    \begin{tabular}{cccc}
        % ================= Example 1 =================
        \begin{minipage}[t]{0.23\textwidth}
            \centering
            \includegraphics[width=0.48\linewidth]{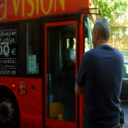}
            \hfill
            \includegraphics[width=0.48\linewidth]{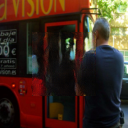}

            \vspace{0.4em}
            {\small
            \textbf{Orig:} \textcolor{green!60!black}{a man} in a blue shirt \\
            \textbf{Inp:} \textcolor{red}{a woman} in a blue shirt \\
            }
        \end{minipage}
        &
        % ================= Example 2 =================
        \begin{minipage}[t]{0.23\textwidth}
            \centering
            \includegraphics[width=0.48\linewidth]{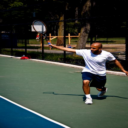}
            \hfill
            \includegraphics[width=0.48\linewidth]{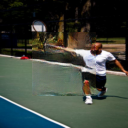}

            \vspace{0.4em}
            {\small
            \textbf{Orig:} a man playing tennis in \textcolor{green!60!black}{blue shorts} \\
            \textbf{Inp:} a man in a white shirt and \textcolor{red}{black shorts} \\
            }
        \end{minipage}
        &
        % ================= Example 3 =================
        \begin{minipage}[t]{0.23\textwidth}
            \centering
            \includegraphics[width=0.48\linewidth]{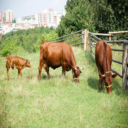}
            \hfill
            \includegraphics[width=0.48\linewidth]{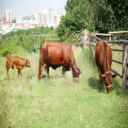}

            \vspace{0.4em}
            {\small
            \textbf{Orig:} \textcolor{green!60!black}{a brown cow} in the middle \\
            \textbf{Inp:} \textcolor{red}{the horse} in the middle \\
            }
        \end{minipage}
        &
        % ================= Example 4 =================
        \begin{minipage}[t]{0.23\textwidth}
            \centering
            \includegraphics[width=0.48\linewidth]{figs/original_1034_124622_158004.png}
            \hfill
            \includegraphics[width=0.48\linewidth]{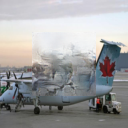}

            \vspace{0.4em}
            {\small
            \textbf{Orig:} \textcolor{green!60!black}{a blue and white plane} \\
            \textbf{Inp:} \textcolor{red}{a blue truck} with a tail \\
            }
        \end{minipage}
    \end{tabular}

    \caption{
    Qualitative examples illustrating captioning errors induced by center-region inpainting.
    Incorrect semantic attributes introduced by inpainting are highlighted in \textcolor{red}{red}, while the correct interpretation is shown in \textcolor{green!60!black}{green}.
    }
    \label{fig:qualitative_inpainting_sidebyside}
\end{figure*}

Vision-language models (VLMs) are increasingly deployed within multi-stage pipelines, where visual inputs are processed or reconstructed before being consumed by language models.
A common instance of this paradigm is image inpainting, in which missing or corrupted regions are filled prior to downstream generative tasks.
Although modern diffusion-based inpainting models produce visually plausible reconstructions, they are optimized primarily for pixel-level realism, allowing subtle but semantically meaningful artifacts to be introduced without being perceptually salient. Figure~\ref{fig:qualitative_inpainting_sidebyside} shows representative examples of this effect.
Despite visually coherent reconstructions, localized inpainting artifacts lead to object substitutions, attribute changes, or category-level errors in downstream captions.

Because captioning models lack explicit awareness of which regions have been reconstructed, synthesized content may be treated as genuine visual evidence.
This raises a central question: \emph{to what extent does reconstruction fidelity influence downstream caption correctness and semantic grounding?}
Despite the widespread use of inpainting and captioning, this relationship remains underexplored. 

To study this interaction, we introduce a two-stage diagnostic framework.
Images are synthetically degraded and reconstructed using diffusion-based inpainting, and both original and reconstructed images are passed to a frozen captioning model.
By directly comparing the resulting captions, this enables controlled analysis of how reconstruction artifacts affect language generation without retraining either model.

Using this framework, we conduct a systematic analysis across diverse domains, examining relationships between reconstruction fidelity metrics and caption quality.
We further analyze representation-level and attention-level changes within frozen vision encoders to understand how reconstruction artifacts manifest internally.
Across datasets, we observe consistent associations between reconstruction fidelity and caption grounding, even when reconstruction and captioning models are trained independently.

\paragraph{Contributions.}
We (i) introduce a model-agnostic diagnostic framework for analyzing reconstruction-induced effects in vision-language pipelines,
(ii) provide multi-dataset empirical evidence linking reconstruction fidelity to caption quality, and
(iii) show that inpainting artifacts are associated with layer-wise and spatially localized changes in vision encoders.

\section{Methodology}

\begin{figure}[h]
    \centering
    \includegraphics[width=\columnwidth, trim={0cm 31cm 0cm 0cm}, clip]{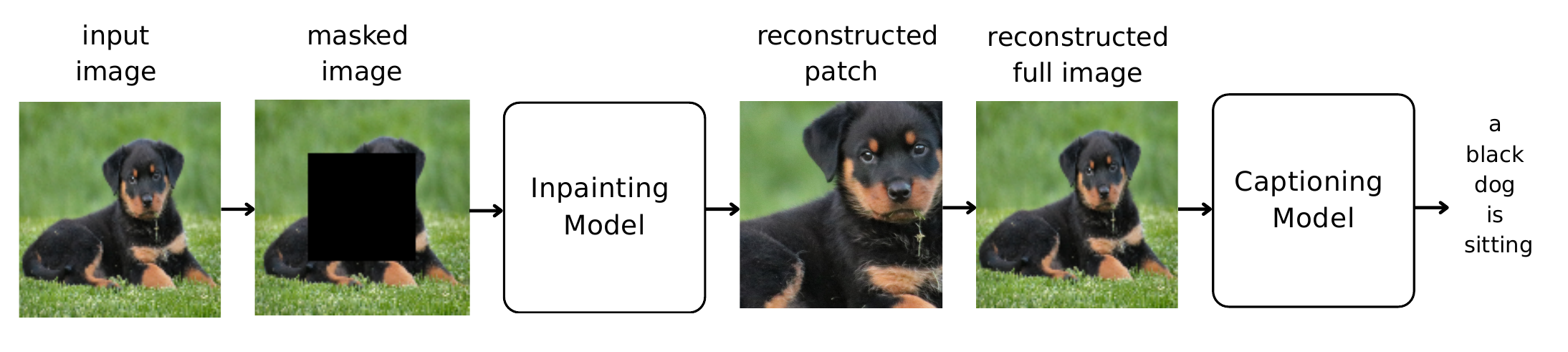}
    \caption{Degradation–reconstruction–captioning framework used to evaluate how inpainting artifacts propagate into downstream language outputs.}
    \label{fig:arch}
\end{figure}

We adopt a degradation–reconstruction–captioning framework as shown in Figure~\ref{fig:arch}. 
Given an input image, we apply a synthetic degradation to a predefined region using perturbations. 
The degraded image is reconstructed with a diffusion-based inpainting model, yielding a visually plausible but potentially semantically altered input.
Both original and reconstructed images are then passed to a frozen captioning model, and the resulting captions are compared using standard linguistic and semantic metrics.

Our analysis proceeds along two complementary axes.
First, we examine correlations between reconstruction fidelity metrics and caption quality metrics to assess whether improvements in visual fidelity correspond to stronger linguistic grounding.
Second, we analyze representational stability within a frozen vision encoder by measuring embedding similarity and layer-wise attention drift between original and reconstructed inputs. By varying only the reconstruction process while keeping all models fixed, this setup isolates the effect of visual artifacts on downstream language behavior.

\section{Experiments}

\begin{figure}[h]
    \centering
    \includegraphics[width=\columnwidth, trim={0cm 22cm 0cm 0cm}, clip]{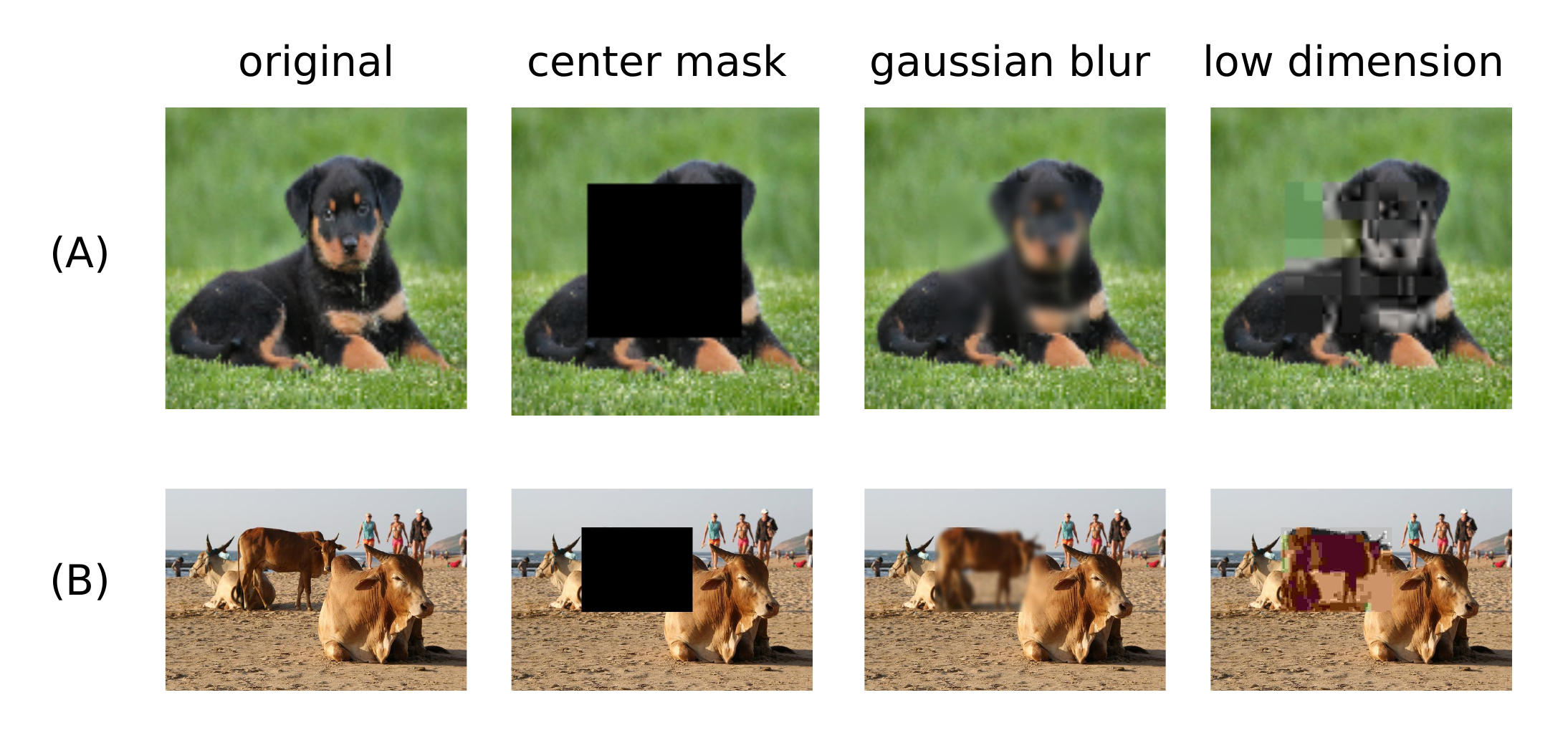}
    \caption{Masking examples illustrating center-region degradations on (A) Flickr and (B) RefCOCOg.}
    \label{fig:maskingegs}
\end{figure}

\begin{figure*}[ht]
    \centering
    \includegraphics[width=\linewidth, trim={0 100 0 0}, clip]{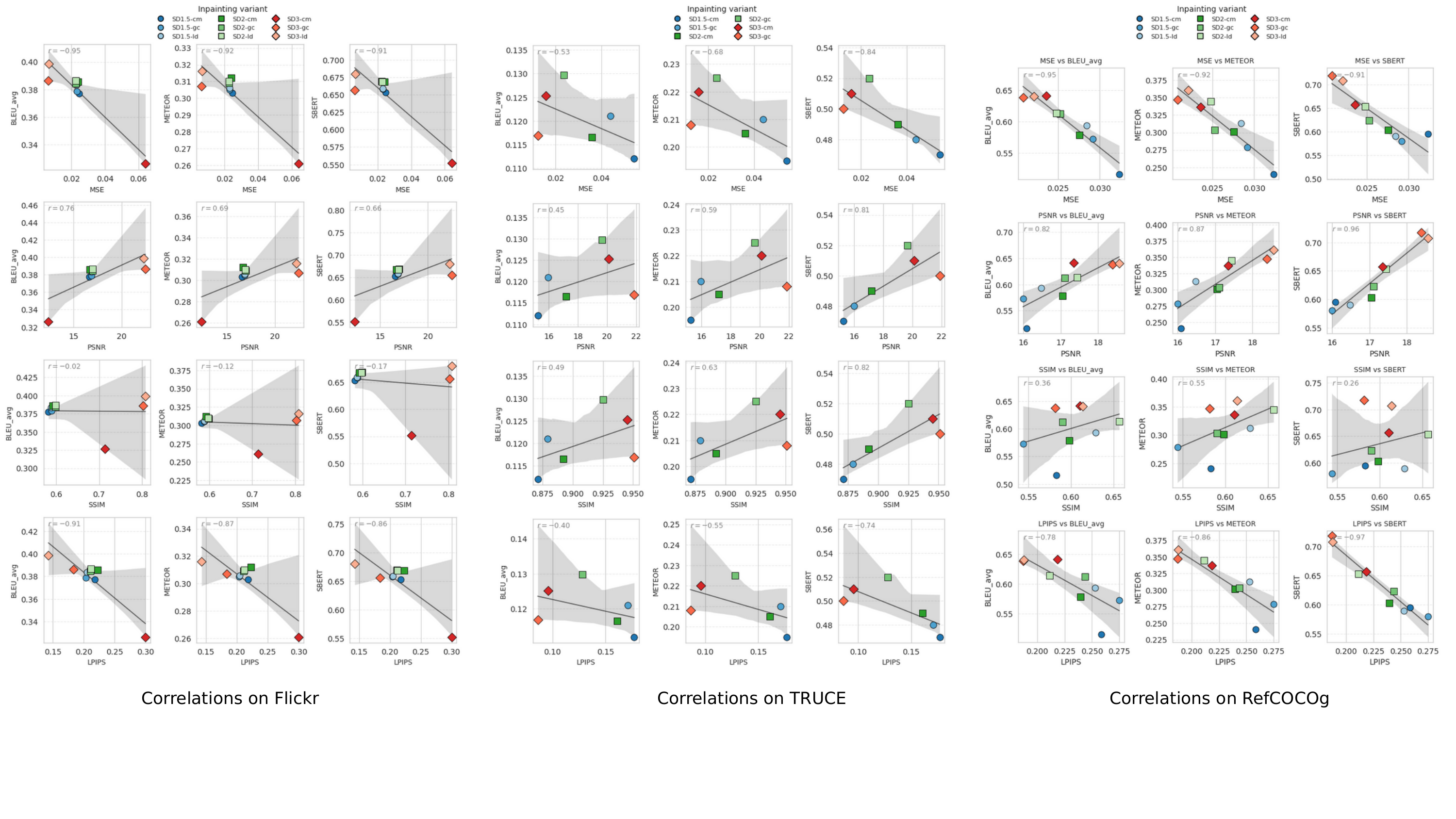}
\caption{
Correlations between reconstruction fidelity metrics and caption quality metrics on Flickr, RefCOCOg, and TRUCE.
Points correspond to Stable Diffusion inpainting variants under three masking strategies:
\emph{cm} (hard center mask), \emph{gc} (Gaussian-blurred center), and \emph{ld} (low-dimensional center degradation).
Caption quality is evaluated using BLIP for Flickr, and Qwen2.5-VL for RefCOCOg and TRUCE.
}
    \label{fig:3domains}
\end{figure*}

\paragraph{Models.} For the inpainting stage, we employ diffusion-based models Stable Diffusion (SD) 1.5, 2.0, and 3.0~\cite{rombach2022high} for masked image completion. These models generate high-fidelity reconstructions while differing in scale and training data diversity which helps us to analyze how architectural and data variations affect downstream robustness.  
We evaluate three vision-language models used for caption generation: LLaVA~\cite{li2024llavaonevisioneasyvisualtask}, BLIP~\cite{li2022blip}, and Qwen2.5-VL~\cite{qwen25vl2024}.
To study visual representations independently of decoding, we extract embeddings and attention maps from a ViT-Base encoder~\cite{dosovitskiy2020vit}.

\paragraph{Evaluation Metrics.}
Reconstruction fidelity is measured using mean squared error (MSE), peak signal-to-noise ratio (PSNR), structural similarity index (SSIM)~\cite{wang2004image}, and learned perceptual image patch similarity (LPIPS)~\cite{zhang2018unreasonable}.
Caption quality is evaluated using BLEU~\cite{papineni2002bleu}, METEOR~\cite{banerjee2005meteor}, ROUGE-L~\cite{lin2004rouge}, and semantic similarity metrics including SimCSE~\cite{gao2022simcsesimplecontrastivelearning} and SBERT.
We further assess representational stability using cosine similarity between visual embeddings and layer-wise attention drift.
All models remain frozen to isolate the impact of reconstruction artifacts.
Full metric definitions and inference settings are provided in Appendix~\ref{sec:metrics} and Appendix~\ref{sec:settings}.

\paragraph{Datasets and Setup.}
We evaluate our framework across multiple datasets, including natural images (Flickr, RefCOCOg), medical imagery (ROCOv2, Indiana X-Ray), audio-spectrograms (GTZAN), and structured time-series plots (TRUCE); representative examples for each dataset are shown in Appendix~\ref{sec:datasets}.
Images are degraded using hard center masking, Gaussian-blurred masking, or low-dimensional compression, reconstructed via diffusion-based inpainting, and then passed to frozen captioning models.
Dataset-specific masking details are provided in Appendix~\ref{sec:masking1}. Figure~\ref{fig:maskingegs} shows how the original image is masked across the three degradation variants.

\section{Results}

\paragraph{Reconstruction fidelity correlates with caption quality.}
Figure~\ref{fig:3domains} reports correlations between reconstruction fidelity metrics and caption quality metrics across SD variants and masking strategies.
Across visually grounded datasets, lower reconstruction error, reflected by lower MSE and LPIPS and higher PSNR, is consistently associated with stronger linguistic alignment.

Among reconstruction metrics, perceptual distance (LPIPS) and pixel-level error (MSE) exhibit the strongest and most consistent correlations with caption quality, indicating that perceptual realism is critical for downstream grounding.
In contrast, SSIM shows weak or inconsistent correlations for natural images such as Flickr, suggesting that global structural similarity alone is insufficient to predict caption correctness.

RefCOCOg and TRUCE follow the same directional trends, with reconstruction fidelity remaining predictive of caption quality.
In RefCOCOg, correlations are particularly strong, reflecting the importance of preserving localized visual content for region-grounded captions.
TRUCE exhibits slightly weaker but still consistent correlations, likely due to the numeric and trend-focused nature of captions, which reduces sensitivity to fine-grained visual detail.

Appendix~\ref{sec:success_case_roco} shows analysis on ROCOv2, showing how reconstruction fidelity relates to caption quality in medical imagery, while Appendix~\ref{sec:failure_cases} documents failure cases (GTZAN and X-ray) where limited linguistic variability prevents meaningful reconstruction–caption correlations.
Across these analyses, ROCOv2 preserves strong, monotonic reconstruction–caption relationships despite higher absolute reconstruction error, whereas GTZAN and X-ray show near-flat caption metrics that remain insensitive to large reconstruction differences.
Additional Flickr analyses, including leave-one-out stability and guidance scale sensitivity, are also reported in Appendix~\ref{app:correlation-stability}.

\paragraph{Masking strategy affects semantic stability.}
Beyond aggregate correlations, masking strategy plays a critical role in semantic stability. We see that smoother degradation schemes such as Gaussian-center masking and low-dimensional compression preserve caption quality more effectively than hard center masking. While center-masked reconstructions often appear visually plausible, they induce larger drops in both lexical and semantic metrics, indicating that abrupt spatial discontinuities disrupt object identity and relational cues relied upon by captioning models.

\begin{figure}[h]
    \centering
    \includegraphics[width=\linewidth, trim={0 250 0 0}, clip]{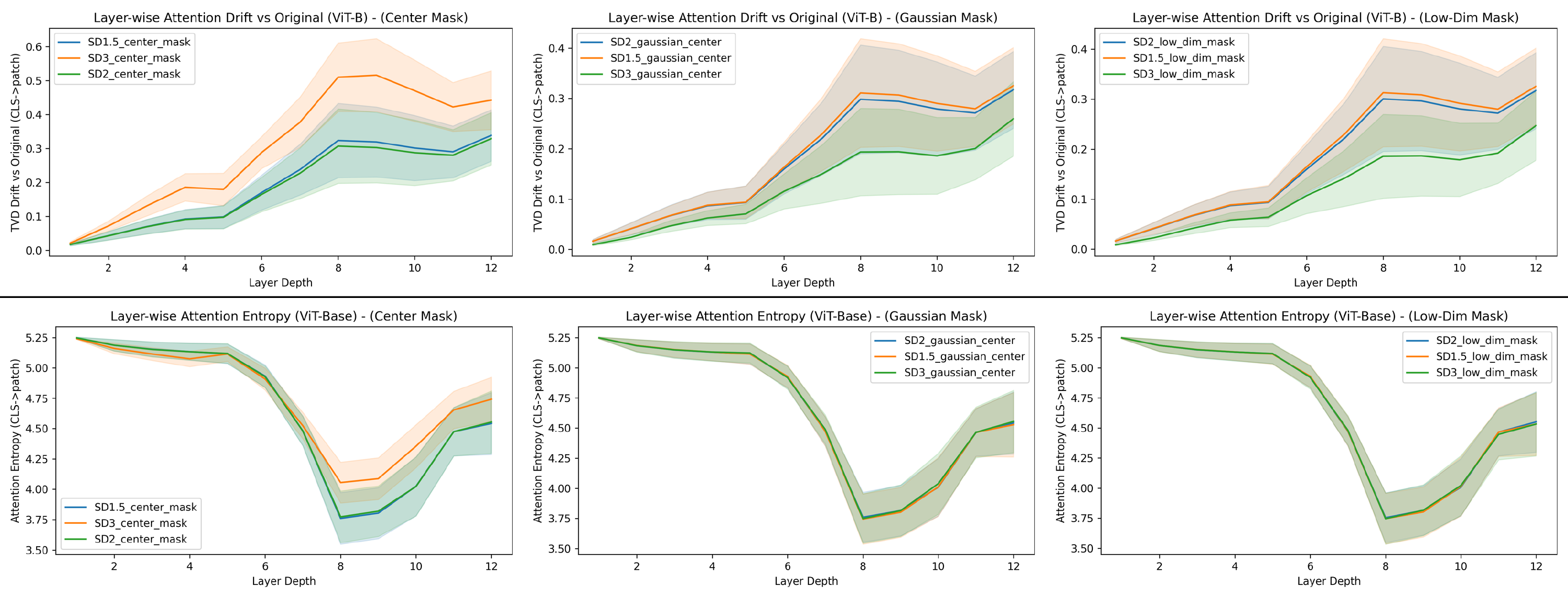}
    \caption{
    \textbf{Layer-wise attention drift and entropy under inpainting on Flickr.}
    Drift increases with depth and is higher for center-masked reconstructions.
    }
    \label{fig:drfts}
\end{figure}

\paragraph{Inpainting induces layer-dependent attention drift.}
We analyze attention patterns in a frozen ViT-Base encoder to assess how reconstruction artifacts affect internal representations.
Figure~\ref{fig:drfts} shows layer-wise Total Variation Distance (TVD) between CLS-to-patch attention maps for original and reconstructed inputs.
Across all inpainting variants, attention drift increases with depth and peaks in later layers, while earlier layers remain comparatively stable.
Center-masked reconstructions consistently induce the largest drift, indicating that sharper visual disruptions have a stronger impact on higher-level representations.
Attention entropy exhibits a complementary pattern, with a dip in mid-layers followed by increased dispersion in deeper layers as semantic features are formed.

\begin{figure}[h]
    \centering
    \includegraphics[width=0.9\linewidth, trim={0 300 0 140}, clip]{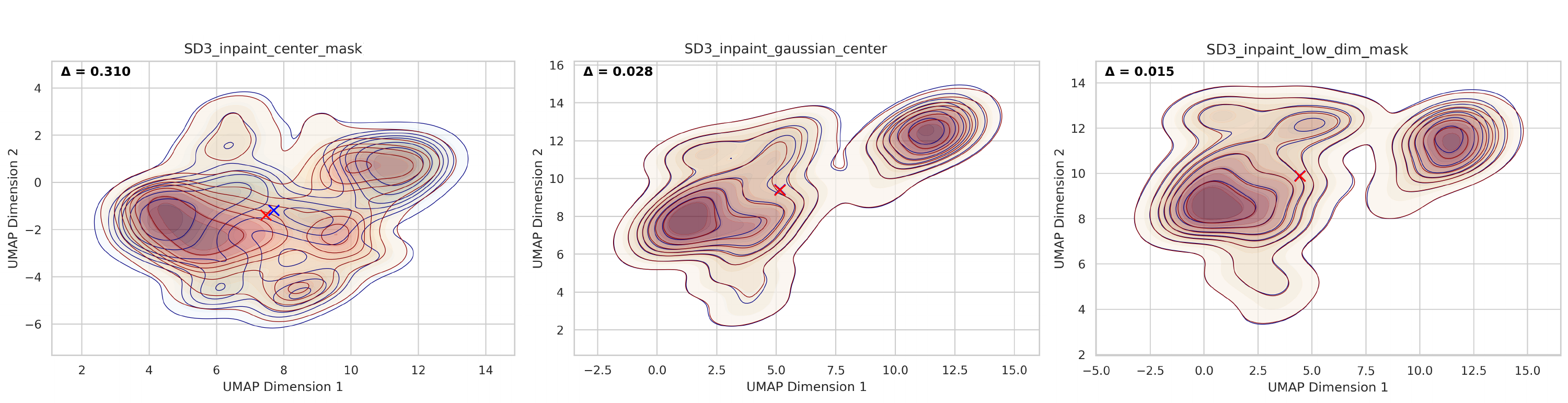}
    \caption{
    \textbf{UMAP density visualization of ViT CLS embeddings for SD3 inpainting variants on Flickr.}
    Smoother masking strategies preserve closer alignment with original representations.
    }
    \label{fig:kde_sd3}
\end{figure}

\begin{figure}[ht]
    \centering
    \includegraphics[width=\linewidth, trim={0 240 0 170}, clip]{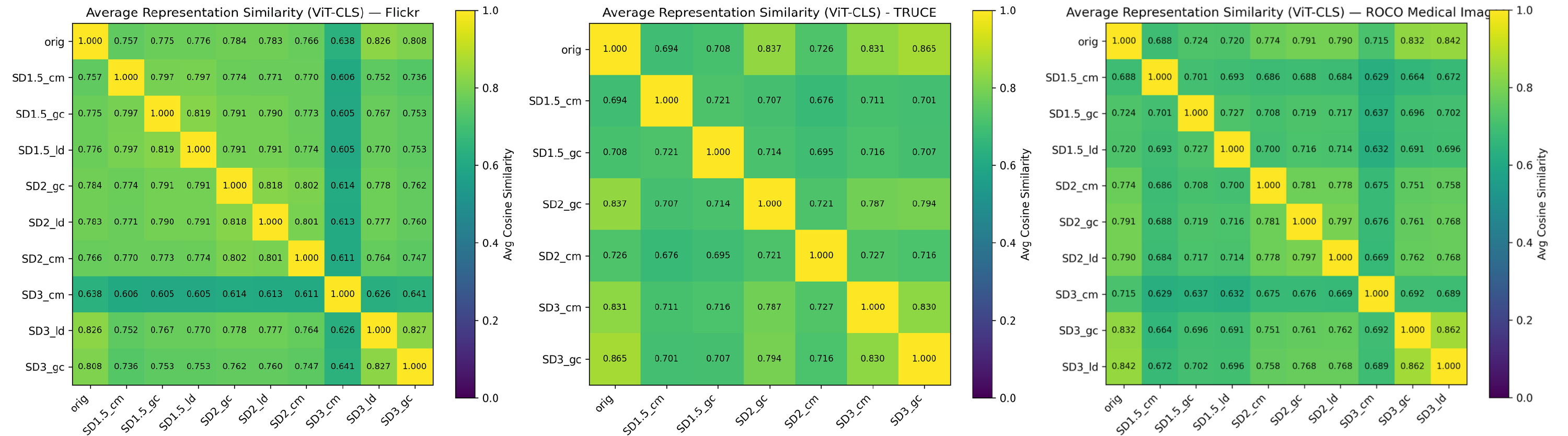}
    \caption{
    Cosine similarity between original and reconstructed visual representations across domains.
    }
    \label{fig:cosine}
\end{figure}

\paragraph{Global visual representations shift under inpainting.}
We further examine how inpainting affects global visual representations. Figure~\ref{fig:kde_sd3} visualizes UMAP projections of ViT CLS embeddings for original and reconstructed images under different SD3 masking strategies on Flickr. Center-masked reconstructions exhibit clear separation from the original embedding distribution, while Gaussian-center and low-dimensional masking yield progressively closer alignment. These trends are consistent with cosine similarity measurements across domains (Figure~\ref{fig:cosine}), where smoother strategies preserve higher representational similarity.

\begin{figure}[ht]
    \centering
    \includegraphics[width=\columnwidth, trim={0 780 0 0}, clip]{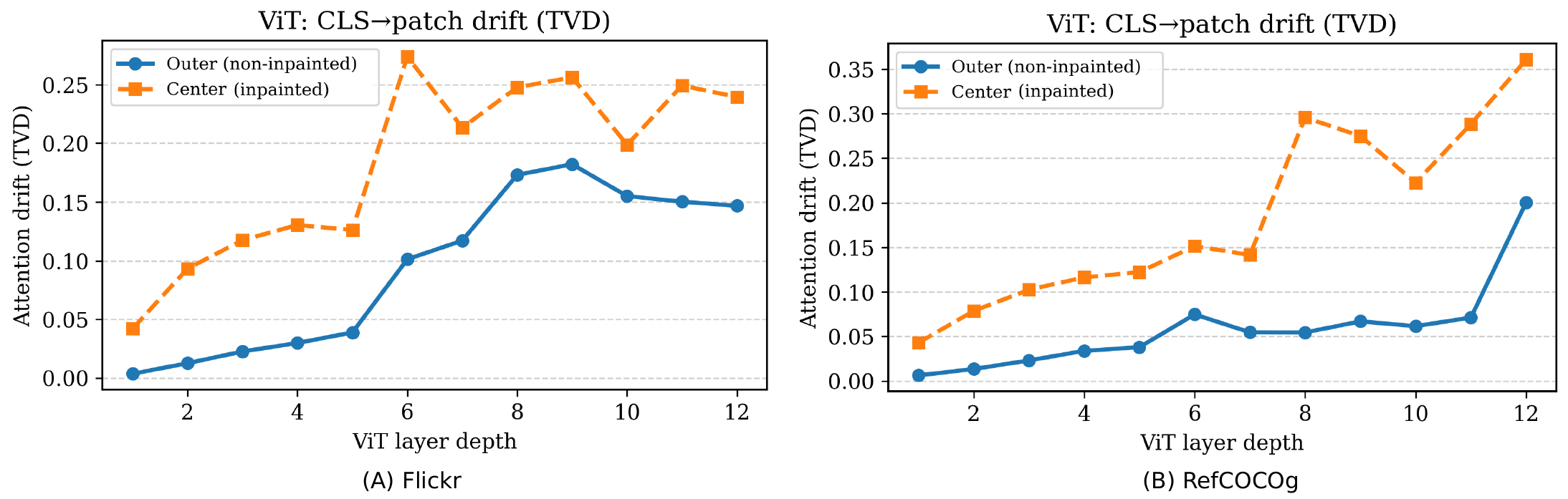}
    \caption{
    Spatial localization of attention drift under center inpainting.
    Drift is consistently higher in reconstructed regions and increases with depth.
    }
    \label{fig:spatial_drift}
\end{figure}

\paragraph{Attention drift is spatially localized.}
Finally, we assess whether attention drift is localized to reconstructed regions. Figure~\ref{fig:spatial_drift} compares CLS-to-patch attention drift for the inpainted center region versus the unmasked outer region. Across all layers, drift is consistently higher within the reconstructed center, with divergence increasing in deeper layers, while outer regions remain comparatively stable. This confirms that semantic instability is aligned with reconstructed content rather than global image degradation.

\section{Conclusion}

We investigated how artifacts introduced by diffusion-based inpainting influence downstream captioning in VLMs.
Across multiple domains with visually grounded captions, improved reconstruction fidelity is consistently associated with more stable captions and visual representations.
Pixel-level and perceptual reconstruction metrics are strong predictors of caption quality, whereas global structural similarity alone is often insufficient.
We further show that inpainting artifacts primarily affect deeper layers of vision encoders and induce attention drift that is spatially aligned with reconstructed regions.
Overall, our findings indicate that inpainting can meaningfully alter semantic processing in multimodal pipelines and motivate reconstruction-aware diagnostics when evaluating vision-language robustness.

\section*{Limitations}

Our study focuses on diffusion-based inpainting as a representative reconstruction mechanism, and the findings may not directly generalize to other visual preprocessing operations such as super-resolution or denoising. While we evaluate multiple Stable Diffusion variants and masking strategies, we do not exhaustively explore reconstruction hyperparameters or alternative architectures, which may influence the degree of reconstruction-induced drift. Our analysis is centered on captioning and visually grounded language generation. The results may not fully extend to other vision-language tasks such as visual question answering or multi-step reasoning. Additionally, some datasets with limited linguistic variability, such as GTZAN and X-ray imagery, restrict the strength of measurable reconstruction-language correlations. Finally, all models are evaluated in a frozen setting to isolate reconstruction effects. This does not capture potential adaptation that may occur in end-to-end trained multimodal systems, which remains an open direction for future work.

\section*{Ethical Considerations}
\label{ethical}

All datasets used in this work are publicly available benchmarks and do not contain personally identifiable information. The study does not involve human subjects, user interaction, or the collection of new personal data. All models are used in an inference-only setting and are not deployed in real-world decision-making contexts. Diffusion-based inpainting can introduce visually plausible but semantically incorrect content. By explicitly analyzing and documenting these failure modes, our work aims to support more transparent and reconstruction-aware evaluation of vision-language systems, particularly in settings where robustness is critical.

\section*{LLM Usage Statement}

Large Language Models were used in a limited and well-defined manner in this work. VLMs were used as evaluation models, serving as frozen captioning baselines. LLMs were used as a writing assistance tool to improve clarity and presentation. They did not contribute to research ideation, experimental design, or analysis. All conclusions and responsibility for the content rest solely with the authors.

% \clearpage
\bibliography{references}
% \printbibliography

\appendix

% \section{Example Appendix}
\label{sec:appendix}

\section*{Appendix}

\section{Related Work}

Early image inpainting methods framed reconstruction as a self-supervised learning problem, with GAN-based approaches such as Context Encoders \cite{pathak2016context} and U-Net–style architectures \cite{ronneberger2015unet} emphasizing spatial coherence and perceptual realism. Subsequent work introduced perceptual similarity metrics \cite{zhang2018unreasonable} and large-scale generative models, including recent diffusion-based inpainting systems that produce visually plausible completions. These models are typically evaluated using pixel-level or perceptual fidelity metrics, without explicit assessment of their semantic reliability when integrated into downstream generation pipelines.

Image captioning has evolved around encoder–decoder architectures with attention mechanisms that condition language generation on visual representations \cite{xu2015show,anderson2018bottom}. Large-scale pretraining approaches such as BLIP \cite{li2022blip} further strengthened vision–language alignment, but captioning models remain sensitive to visual input quality and grounding cues.

A related line of work studies object hallucination in image captioning, where models generate entities not present in the image. Rohrbach et al.~\cite{rohrbach2019objecthallucinationimagecaptioning} showed that strong performance on standard captioning metrics does not necessarily guarantee faithful visual grounding, as captioning models may rely on language priors even when visual evidence is ambiguous or weak. More recent studies extend this analysis to large VLMs, demonstrating that hallucination remains prevalent even in LLM-based systems \cite{liu2024surveyhallucinationlargevisionlanguage}. These works primarily focus on model architectures, decoding behavior, and instruction design, rather than on the effects of upstream visual transformations.

Separately, robustness benchmarks such as ImageNet-C and ImageNet-P \cite{hendrycks2019benchmarkingneuralnetworkrobustness} evaluate classifier stability under common corruptions and perturbations. While influential for vision robustness, these benchmarks do not consider reconstruction-based transformations or their interaction with downstream language generation.

Our work complements these lines of research by analyzing how reconstruction artifacts introduced by diffusion-based inpainting propagate into caption grounding and internal visual representations in frozen vision–language pipelines.

\section{Datasets}
\label{sec:datasets}

\subsection{Dataset Descriptions}

Below, we summarize the datasets used and the subsets selected for our experiments.

\begin{figure}[ht]
    \centering
    \includegraphics[width=\columnwidth, trim={0 0 380 0}, clip]
    {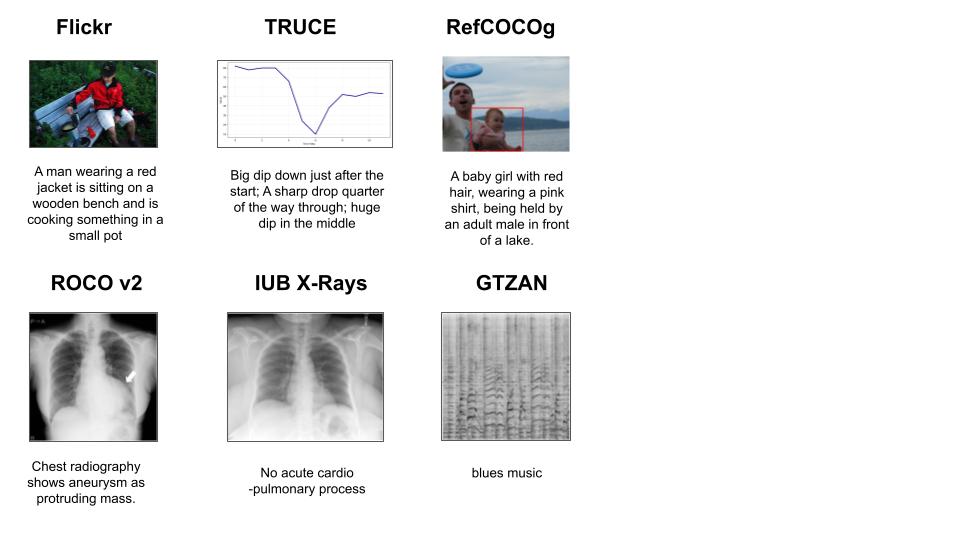}
    \caption{Representative examples from each dataset used in this study. These examples illustrate the diversity of visual modalities and caption styles across datasets.}
    \label{fig:dataset_examples}
\end{figure}

\paragraph{Flickr Entities.}
Flickr Entities~\cite{plummer2016flickr30kentitiescollectingregiontophrase} pairs natural images with region-level annotations and human-written captions. We select a subset of 4,000 image-caption pairs for our experiments. Captions are descriptive, and visually grounded.

\paragraph{RefCOCOg.}
RefCOCOg~\cite{yu2016modelingcontextreferringexpressions} provides referring expressions aligned with explicit bounding box annotations. Each image is associated with one or more localized descriptions targeting specific objects or regions. We use a split of 2,000 samples and apply masking within the annotated bounding boxes, enabling evaluation of reconstruction effects under spatial grounding.

\paragraph{Medical Imaging (Indiana Chest X-ray and ROCOv2).}
We use two medical image captioning datasets to evaluate reconstruction effects in high-stakes domains. The Indiana University Chest X-ray dataset~\cite{demner-fushman2016preparing} pairs chest radiographs with radiology reports; we extract the \emph{Impression} section as the target caption. ROCOv2 \cite{Rckert2024ROCOv2RO} extends this setting with medically curated image-caption pairs and structured clinical concepts. For both datasets, we use a randomly sampled subset of 1,000 images. The average caption length across medical datasets is approximately 149 characters.

\paragraph{GTZAN Spectrograms.}
GTZAN \cite{tzanetakis2002musical} is an audio dataset containing music recordings from 10 genre categories. We uniformly sample 500 audio files (50 per genre), resampled to 44.1\,kHz. Each clip is truncated to the first 5.12\,s and converted into a mel-spectrogram, which is treated as a 2D image. Genre labels serve as short textual descriptors.

\paragraph{TRUCE Time-Series Plots.}
TRUCE \cite{jhamtani2021truthconditionalcaptioningtimeseries} consists of numeric time series paired with textual captions describing temporal properties such as trends, peaks, and anomalies. We render each series as a line plot prior to inpainting and captioning.

\subsection{Masking and Degradation Strategies}
\label{sec:masking1}

To study how different forms of visual degradation affect downstream captioning and representation stability, we apply three controlled masking strategies to the input images prior to reconstruction. All masking operations are applied deterministically and only affect a localized region of the input, while the remainder of the image is left unchanged.

\paragraph{Masking Variants.}
We consider the following three degradation types:

\begin{itemize}[leftmargin=*]
    \item \textbf{Center Mask.}
    A rectangular region covering the target area is fully removed by setting all pixel values to zero. This produces a sharp spatial discontinuity and removes all visual information within the masked region.

    \item \textbf{Gaussian Blur.}
    The target region is degraded using a Gaussian blur with a fixed kernel size. Each pixel is replaced by a weighted average of its neighbors, resulting in a smooth attenuation of high-frequency details while preserving coarse structure.

    \item \textbf{Low-Dimensional Center Degradation.}
    Instead of fully removing the region, we apply an aggressive but structured degradation that preserves spatial layout while eliminating fine-grained semantic cues. Specifically, the masked region undergoes:
    (i) color quantization via $k$-means clustering with $k{=}4$ colors,
    (ii) spatial downsampling followed by upsampling to suppress high-frequency texture, and
    (iii) extremely low-quality JPEG compression.
    This produces a visually coherent patch that retains approximate shape and layout but loses texture, color fidelity, and detailed semantics.
\end{itemize}

\paragraph{Dataset-Specific Masking.}
\label{sec:masking2}
For Flickr, medical image datasets and audio dataset, masking is applied to a fixed central region of the image. In contrast, RefCOCOg provides localized referring expressions paired with bounding boxes. For this dataset, we apply all masking operations directly within the annotated bounding box corresponding to the target object or region. As a result, the masked area may occur anywhere in the image and is semantically aligned with the referring expression. For TRUCE time-series plots, masking is applied along the temporal axis rather than a fixed spatial location. We select one of several informative contiguous segments of the plotted curve and degrade approximately 25\% of the series length. This ensures that the degraded region corresponds to a meaningful portion of the temporal dynamics while preserving the overall structure of the plot. Gaussian blur and low-dimensional degradation use the same hyperparameters as in the natural image setting.

\section{Metrics}
\label{sec:metrics}

We evaluate the impact of inpainting artifacts using complementary metrics that quantify
(i) reconstruction fidelity,
(ii) caption quality, and
(iii) representation and attention stability.
All metrics compare outputs from reconstructed inputs against their original counterparts.

\paragraph{Reconstruction Fidelity.}
Reconstruction quality is evaluated within the degraded regions using a combination of pixel-level, signal-level, and perceptual metrics. We report MSE to quantify pixel-wise reconstruction error and PSNR to measure signal fidelity relative to reconstruction noise. Structural consistency between the original and reconstructed regions is assessed using the SSIM \cite{wang2004image}. To capture perceptual differences beyond low-level statistics, we additionally report LPIPS \cite{zhang2018unreasonable}, which measures distance in deep feature embedding space.

\paragraph{Caption Quality.}
Generated captions are evaluated against ground truths using standard lexical and semantic metrics. Lexical overlap is measured using BLEU-1 through BLEU-4 \cite{papineni2002bleu}, which capture $n$-gram precision at increasing orders, and ROUGE-L \cite{lin2004rouge}, which evaluates longest common subsequence recall. We also report METEOR \cite{banerjee2005meteor}, an alignment-based metric that incorporates synonymy and stemming to better reflect semantic similarity. To assess semantic alignment beyond surface overlap, we additionally compute cosine similarity between sentence embeddings using SBERT~\cite{reimers2019sentencebertsentenceembeddingsusing} and both supervised and unsupervised variants of SimCSE~\cite{gao2022simcsesimplecontrastivelearning}. For each generated caption, we report the maximum similarity to any reference caption for the corresponding input.

\paragraph{Representation Similarity.}
To quantify global visual drift, we extract CLS embeddings from a frozen ViT encoder and compute cosine similarity between embeddings obtained from original and reconstructed images.

\paragraph{Attention Drift and Entropy.}
To analyze how inpainting artifacts propagate through model internals, we quantify changes in attention behavior using divergence-based metrics computed from CLS-to-patch attention maps. We measure total variation distance to capture layer-wise divergence between attention distributions obtained from original and reconstructed inputs. In addition, we compute attention entropy to characterize the dispersion of CLS-to-patch attention within each layer, providing a measure of how concentrated or diffuse the model’s visual focus becomes following reconstruction. TVD quantifies divergence between attention distributions as the sum of absolute differences across patch positions. For spatial analyses, TVD is computed separately over inpainted and non-inpainted regions using binary mask supervision.

\section{Settings and Configurations}
\label{sec:settings}

All experiments are conducted under a unified inference-only evaluation protocol. Across all datasets, models are used in a frozen state with no fine-tuning or adaptation, ensuring that observed effects arise solely from input degradation and reconstruction rather than model updates. 

\paragraph{Captioning Settings.}
Caption generation follows a shared decoding configuration across models, using beam search with six beams, top-$p$ sampling with $p=0.9$, temperature $0.8$, and a maximum of $48$-$64$ generated tokens depending on the task. For each image, three candidate captions are generated. Prompt conditioning is used where supported: RefCOCOg captions are explicitly constrained to describe the contents of the red bounding box, while BLIP operates in an unconditional captioning mode due to limited prompt adherence.

\paragraph{Diffusion-Based Inpainting.}
All models are applied using fixed hyperparameters across datasets. 
Inpainting is conducted with $50$ inference steps in all cases. 
Guidance scales are fixed within the empirically stable value of $7.5$ identified in preliminary analysis, and are held constant per diffusion model across all experiments. 
The strength parameter is set to $1.0$ for SD1.5 and SD2. For SD3, strength is reduced to $0.6$, as higher values were observed to either overwrite large portions of the original image or collapse to degenerate reconstructions in which the masked region remains blank.

\paragraph{Prompt Construction for Inpainting.}
Inpainting prompts are dataset-specific but strictly derived from existing annotations. For Flickr, captions associated with the original images are used directly as prompts. For RefCOCOg, multiple bounding-box captions are concatenated into a single prompt describing the target region, truncated to a maximum of $75$ tokens to avoid prompt overflow. For TRUCE plots, prompts consist of a short reconstruction instruction followed by the raw numeric time-series and its description.

\section{Success Cases}
\label{sec:success_cases}

\subsection{Flickr dataset (Success Case)}
\label{sec:success_case_flickr}

\paragraph{Analysis}

Flickr exhibits a clear and stable reconstruction–caption relationship.
MSE and LPIPS maintain strong negative correlations with all caption metrics across inpainting variants, while PSNR remains positively correlated.
SSIM shows weak and variable behavior, reinforcing that structural similarity alone is not a reliable predictor of caption grounding.
Gaussian-blurred and low-dimensional masking under SD2 and SD3 yield the most stable trade-offs between reconstruction fidelity and caption quality, whereas hard center masking introduces disproportionate degradation due to removal of salient semantic content (see Table~\ref{tab:recon_metrics_flickr} and~\ref{tab:og_with_percent}).

\paragraph{Correlation Robustness Results on Flickr}
\label{app:correlation-stability}

We assess the robustness of reconstruction-caption correlations on Flickr using a leave-one-out (LOO) analysis, recomputing Pearson correlations after removing each inpainting configuration in turn. Table~\ref{tab:loo_correlation_flickr} summarizes the stability of each reconstruction metric across all caption quality measures. 

MSE shows consistently strong negative correlations with all caption metrics, with full-data correlations in the range $[-0.951, -0.906]$ and no sign reversals under leave-one-out (LOO) analysis. LPIPS follows a similar pattern, exhibiting strong negative correlations, low LOO variance, and stable signs across splits. PSNR displays moderate positive correlations with caption quality; while its magnitude varies more across LOO splits, the correlation direction remains consistent. In contrast, SSIM exhibits weak and unstable behavior, with near-zero full correlations and frequent sign reversals under LOO analysis, indicating limited reliability for predicting semantic caption quality.

\begin{table}[ht]
\centering
\scriptsize
\setlength{\tabcolsep}{6pt}
\renewcommand{\arraystretch}{1.15}
\caption{\textbf{Leave-one-out (LOO) correlation stability on Flickr.}
For each reconstruction metric, we report the range of full-data Pearson correlations with caption metrics, the mean and standard deviation of LOO correlations, and the number of sign reversals across all LOO splits.}
\label{tab:loo_correlation_flickr}
\begin{tabular}{lcccc}
\toprule
\textbf{Metric} & \textbf{$r_{\text{full}}$ range} & $\boldsymbol{\mu_{\text{LOO}}}$ & $\boldsymbol{\sigma_{\text{LOO}}}$ & \textbf{Sign flips} \\
\midrule
MSE   & $[-0.951, -0.906]$ & $\approx -0.88$ & $\leq 0.19$ & 0 \\
LPIPS & $[-0.914, -0.857]$ & $\approx -0.85$ & $\leq 0.11$ & 0 \\
PSNR  & $[0.662, 0.761]$   & $\approx 0.69$  & $\leq 0.13$ & 0 \\
SSIM  & $[-0.166, -0.022]$ & $\approx -0.07$ & $\geq 0.19$ & $\geq 1$ \\
\bottomrule
\end{tabular}
\end{table}

\begin{table}[H]
\centering
\scriptsize
\setlength{\tabcolsep}{3.5pt}
\renewcommand{\arraystretch}{1.15}
\caption{\textbf{Pixel-level reconstruction metrics on Flickr.}
MSE $\downarrow$, PSNR $\uparrow$, SSIM $\uparrow$, and LPIPS $\downarrow$ are averaged over 4k samples.
Lower MSE/LPIPS and higher PSNR/SSIM indicate better reconstruction fidelity.}
\resizebox{0.8\columnwidth}{!}{%
\begin{tabular}{lrrrr}
\toprule
\textbf{Method} &
\textbf{MSE} &
\textbf{PSNR} &
\textbf{SSIM} &
\textbf{LPIPS} \\
\midrule
SD1.5-cm & 0.0246 & 16.59 & 0.582 & 0.218 \\
SD1.5-gc & 0.0231 & \textbf{16.88} & \textbf{0.589} & \textbf{0.204} \\
SD1.5-ld & \textbf{0.0231} & 16.86 & 0.588 & 0.205 \\
\midrule
SD2-cm & 0.0240 & 16.68 & 0.592 & 0.223 \\
SD2-gc & \textbf{0.0225} & \textbf{16.98} & \textbf{0.599} & 0.211 \\
SD2-ld & 0.0226 & 16.94 & 0.598 & \textbf{0.212} \\
\midrule
SD3-cm & 0.0641 & 12.37 & 0.714 & 0.300 \\
SD3-gc & \textbf{0.00629} & \textbf{22.47} & 0.803 & 0.184 \\
SD3-ld & 0.00666 & 22.26 & \textbf{0.808} & \textbf{0.143} \\
\bottomrule
\end{tabular}
}
\label{tab:recon_metrics_flickr}
\end{table}

\begin{table*}[ht]
\centering
\scriptsize
\setlength{\tabcolsep}{2.5pt}
\renewcommand{\arraystretch}{1.15}
\caption{\textbf{Captioning performance (absolute and relative) across inpainting variants and original datasets for Flickr.}
Lexical metrics: BLEU-1–4, METEOR (MET.), ROUGE-L (R-L).  
Semantic metrics: supervised SimCSE (sup.), unsupervised SimCSE (unsup.), SBERT cosine similarity.  
Abbreviations — cm: center mask; gc: Gaussian center; ld: low-dimensional mask.  
$\%\Delta = (\text{Inpainted} - \text{Original})/\text{Original} \times 100$. Negative = performance drop.}
\resizebox{\textwidth}{!}{%
\begin{tabular}{lrrrrrrrrrrrrrrrrrrrr}
\toprule
\textbf{Dataset} & B1 & $\%\Delta$ & B2 & $\%\Delta$ & B3 & $\%\Delta$ & B4 & $\%\Delta$ & MET. & $\%\Delta$ & R-L & $\%\Delta$ & sup. & $\%\Delta$ & unsup. & $\%\Delta$ & SBERT & $\%\Delta$ \\
\midrule
\multicolumn{19}{c}{\textbf{BLIP}} \\
\midrule
SD1.5-cm & 0.602 & \cellcolor{lightred}{-3.68} & 0.423 & \cellcolor{lightred}{-5.79} & 0.289 & \cellcolor{lightred}{-7.67} & 0.195 & \cellcolor{lightred}{-10.14} & 0.303 & \cellcolor{lightred}{-3.81} & 0.453 & \cellcolor{lightred}{-3.63} & 0.753 & \cellcolor{lightred}{-2.09} & 0.716 & \cellcolor{lightred}{-2.16} & 0.653 & \cellcolor{lightred}{-2.83} \\
SD1.5-gc & 0.604 & \cellcolor{lightred}{-3.36} & 0.425 & \cellcolor{lightred}{-5.35} & 0.291 & \cellcolor{lightred}{-7.01} & 0.195 & \cellcolor{lightred}{-10.14} & 0.305 & \cellcolor{lightred}{-3.17} & 0.454 & \cellcolor{lightred}{-3.40} & 0.756 & \cellcolor{lightred}{-1.69} & 0.720 & \cellcolor{lightred}{-1.64} & 0.658 & \cellcolor{lightred}{-2.08} \\
SD1.5-ld & 0.608 & \cellcolor{lightred}{-2.72} & 0.430 & \cellcolor{lightred}{-4.23} & 0.296 & \cellcolor{lightred}{-5.43} & 0.200 & \cellcolor{lightred}{-7.83} & 0.306 & \cellcolor{lightred}{-2.86} & 0.457 & \cellcolor{lightred}{-2.81} & 0.756 & \cellcolor{lightred}{-1.69} & 0.722 & \cellcolor{lightred}{-1.37} & 0.659 & \cellcolor{lightred}{-1.93} \\
SD2-cm   & 0.610 & \cellcolor{lightred}{-2.40} & 0.434 & \cellcolor{lightred}{-3.34} & 0.298 & \cellcolor{lightred}{-4.47} & 0.201 & \cellcolor{lightred}{-7.37} & 0.312 & \cellcolor{lightred}{-0.95} & 0.461 & \cellcolor{lightred}{-1.87} & 0.765 & \cellcolor{lightred}{-0.52} & 0.726 & \cellcolor{lightred}{-0.82} & 0.668 & \cellcolor{lightred}{-0.59} \\
SD2-gc   & 0.610 & \cellcolor{lightred}{-2.40} & 0.431 & \cellcolor{lightred}{-3.99} & 0.296 & \cellcolor{lightred}{-5.43} & 0.201 & \cellcolor{lightred}{-7.37} & 0.309 & \cellcolor{lightred}{-1.90} & 0.460 & \cellcolor{lightred}{-2.13} & 0.764 & \cellcolor{lightred}{-0.65} & 0.727 & \cellcolor{lightred}{-0.69} & 0.669 & \cellcolor{lightred}{-0.45} \\
SD2-ld   & 0.614 & \cellcolor{lightred}{-1.76} & 0.434 & \cellcolor{lightred}{-3.34} & 0.297 & \cellcolor{lightred}{-5.11} & 0.201 & \cellcolor{lightred}{-7.37} & 0.310 & \cellcolor{lightred}{-1.59} & 0.462 & \cellcolor{lightred}{-1.70} & 0.763 & \cellcolor{lightred}{-0.78} & 0.725 & \cellcolor{lightred}{-0.96} & 0.668 & \cellcolor{lightred}{-0.59} \\
SD3-cm   & 0.546 & \cellcolor{lightred}{-12.64} & 0.365 & \cellcolor{lightred}{-18.67} & 0.239 & \cellcolor{lightred}{-23.71} & 0.155 & \cellcolor{lightred}{-28.57} & 0.261 & \cellcolor{lightred}{-17.14} & 0.404 & \cellcolor{lightred}{-14.04} & 0.670 & \cellcolor{lightred}{-12.85} & 0.639 & \cellcolor{lightred}{-12.72} & 0.552 & \cellcolor{lightred}{-17.86} \\
SD3-gc   & 0.608 & \cellcolor{lightred}{-2.72} & 0.433 & \cellcolor{lightred}{-3.56} & 0.299 & \cellcolor{lightred}{-4.47} & 0.205 & \cellcolor{lightred}{-5.53} & 0.307 & \cellcolor{lightred}{-2.54} & 0.456 & \cellcolor{lightred}{-2.98} & 0.757 & \cellcolor{lightred}{-1.56} & 0.719 & \cellcolor{lightred}{-1.78} & 0.656 & \cellcolor{lightred}{-2.38} \\
SD3-ld   & 0.623 & \cellcolor{lightred}{-0.32} & 0.447 & \cellcolor{lightred}{-0.45} & 0.311 & \cellcolor{lightred}{-0.64} & 0.214 & \cellcolor{lightred}{-1.38} & 0.316 & \cellcolor{lightgreen}{+0.32} & 0.468 & \cellcolor{lightred}{-0.43} & 0.774 & \cellcolor{lightgreen}{+0.65} & 0.736 & \cellcolor{lightgreen}{+0.55} & 0.680 & \cellcolor{lightgreen}{+1.19} \\
\emph{Orig.} & 0.625 & — & 0.449 & — & 0.313 & — & 0.217 & — & 0.315 & — & 0.470 & — & 0.769 & — & 0.732 & — & 0.672 & — \\
\midrule
\multicolumn{19}{c}{\textbf{QWEN}} \\
\midrule
SD1.5-cm & 0.555 & \cellcolor{lightred}{-6.25} & 0.386 & \cellcolor{lightred}{-9.59} & 0.263 & \cellcolor{lightred}{-12.63} & 0.178 & \cellcolor{lightred}{-14.83} & 0.322 & \cellcolor{lightred}{-6.67} & 0.452 & \cellcolor{lightred}{-5.98} & 0.781 & \cellcolor{lightred}{-2.98} & 0.726 & \cellcolor{lightred}{-2.94} & 0.674 & \cellcolor{lightred}{-4.12} \\
SD1.5-gc & 0.562 & \cellcolor{lightred}{-5.06} & 0.391 & \cellcolor{lightred}{-8.43} & 0.268 & \cellcolor{lightred}{-10.96} & 0.181 & \cellcolor{lightred}{-13.40} & 0.326 & \cellcolor{lightred}{-5.51} & 0.457 & \cellcolor{lightred}{-4.81} & 0.785 & \cellcolor{lightred}{-2.48} & 0.731 & \cellcolor{lightred}{-2.27} & 0.678 & \cellcolor{lightred}{-3.55} \\
SD1.5-ld & 0.564 & \cellcolor{lightred}{-4.73} & 0.394 & \cellcolor{lightred}{-7.72} & 0.270 & \cellcolor{lightred}{-10.30} & 0.183 & \cellcolor{lightred}{-12.44} & 0.328 & \cellcolor{lightred}{-4.93} & 0.459 & \cellcolor{lightred}{-4.56} & 0.786 & \cellcolor{lightred}{-2.36} & 0.732 & \cellcolor{lightred}{-2.14} & 0.680 & \cellcolor{lightred}{-3.28} \\
SD2-cm   & 0.565 & \cellcolor{lightred}{-4.56} & 0.396 & \cellcolor{lightred}{-7.26} & 0.270 & \cellcolor{lightred}{-10.30} & 0.182 & \cellcolor{lightred}{-12.92} & 0.329 & \cellcolor{lightred}{-4.64} & 0.458 & \cellcolor{lightred}{-4.77} & 0.788 & \cellcolor{lightred}{-2.12} & 0.732 & \cellcolor{lightred}{-2.14} & 0.683 & \cellcolor{lightred}{-2.84} \\
SD2-gc   & 0.575 & \cellcolor{lightred}{-2.87} & 0.405 & \cellcolor{lightred}{-5.15} & 0.281 & \cellcolor{lightred}{-6.64} & 0.193 & \cellcolor{lightred}{-7.66} & 0.332 & \cellcolor{lightred}{-3.77} & 0.465 & \cellcolor{lightred}{-3.32} & 0.791 & \cellcolor{lightred}{-1.74} & 0.736 & \cellcolor{lightred}{-1.60} & 0.688 & \cellcolor{lightred}{-2.13} \\
SD2-ld   & 0.569 & \cellcolor{lightred}{-3.88} & 0.400 & \cellcolor{lightred}{-6.32} & 0.275 & \cellcolor{lightred}{-8.64} & 0.188 & \cellcolor{lightred}{-10.05} & 0.331 & \cellcolor{lightred}{-4.06} & 0.463 & \cellcolor{lightred}{-3.74} & 0.791 & \cellcolor{lightred}{-1.74} & 0.733 & \cellcolor{lightred}{-2.01} & 0.687 & \cellcolor{lightred}{-2.27} \\
SD3-cm   & 0.431 & \cellcolor{lightred}{-27.18} & 0.271 & \cellcolor{lightred}{-36.54} & 0.169 & \cellcolor{lightred}{-43.82} & 0.106 & \cellcolor{lightred}{-49.28} & 0.250 & \cellcolor{lightred}{-27.54} & 0.344 & \cellcolor{lightred}{-28.48} & 0.667 & \cellcolor{lightred}{-17.13} & 0.579 & \cellcolor{lightred}{-22.57} & 0.530 & \cellcolor{lightred}{-24.61} \\
SD3-gc   & 0.561 & \cellcolor{lightred}{-5.23} & 0.396 & \cellcolor{lightred}{-7.26} & 0.274 & \cellcolor{lightred}{-9.03} & 0.189 & \cellcolor{lightred}{-9.57} & 0.334 & \cellcolor{lightred}{-3.19} & 0.457 & \cellcolor{lightred}{-5.00} & 0.789 & \cellcolor{lightred}{-1.99} & 0.729 & \cellcolor{lightred}{-2.54} & 0.682 & \cellcolor{lightred}{-2.99} \\
SD3-ld   & 0.579 & \cellcolor{lightred}{-2.20} & 0.413 & \cellcolor{lightred}{-3.28} & 0.290 & \cellcolor{lightred}{-3.65} & 0.200 & \cellcolor{lightred}{-4.31} & 0.346 & \cellcolor{lightgreen}{+0.29} & 0.469 & \cellcolor{lightred}{-2.50} & 0.803 & \cellcolor{lightred}{-0.25} & 0.746 & \cellcolor{lightred}{-0.27} & 0.703 & \cellcolor{lightgreen}{0.00} \\
\emph{Orig.} & 0.592 & — & 0.427 & — & 0.301 & — & 0.209 & — & 0.345 & — & 0.481 & — & 0.805 & — & 0.748 & — & 0.703 & — \\
\midrule
\multicolumn{19}{c}{\textbf{LLAVA}} \\
\midrule
SD1.5-cm & 0.729 & \cellcolor{lightred}{-6.65} & 0.560 & \cellcolor{lightred}{-9.53} & 0.412 & \cellcolor{lightred}{-12.41} & 0.296 & \cellcolor{lightred}{-15.69} & 0.302 & \cellcolor{lightred}{-10.12} & 0.554 & \cellcolor{lightred}{-6.27} & 0.796 & \cellcolor{lightred}{-3.63} & 0.759 & \cellcolor{lightred}{-3.81} & 0.711 & \cellcolor{lightred}{-4.56} \\
SD1.5-gc & 0.737 & \cellcolor{lightred}{-5.63} & 0.570 & \cellcolor{lightred}{-7.91} & 0.425 & \cellcolor{lightred}{-9.66} & 0.310 & \cellcolor{lightred}{-11.79} & 0.306 & \cellcolor{lightred}{-8.93} & 0.560 & \cellcolor{lightred}{-5.34} & 0.800 & \cellcolor{lightred}{-3.15} & 0.762 & \cellcolor{lightred}{-3.42} & 0.715 & \cellcolor{lightred}{-4.03} \\
SD1.5-ld & 0.737 & \cellcolor{lightred}{-5.63} & 0.570 & \cellcolor{lightred}{-7.91} & 0.426 & \cellcolor{lightred}{-9.55} & 0.312 & \cellcolor{lightred}{-11.12} & 0.305 & \cellcolor{lightred}{-9.23} & 0.560 & \cellcolor{lightred}{-5.34} & 0.801 & \cellcolor{lightred}{-3.03} & 0.764 & \cellcolor{lightred}{-3.17} & 0.715 & \cellcolor{lightred}{-4.03} \\
SD2-cm   & 0.738 & \cellcolor{lightred}{-5.50} & 0.572 & \cellcolor{lightred}{-7.60} & 0.426 & \cellcolor{lightred}{-9.55} & 0.310 & \cellcolor{lightred}{-11.69} & 0.309 & \cellcolor{lightred}{-8.04} & 0.565 & \cellcolor{lightred}{-4.39} & 0.806 & \cellcolor{lightred}{-2.42} & 0.765 & \cellcolor{lightred}{-3.04} & 0.723 & \cellcolor{lightred}{-2.95} \\
SD2-gc   & 0.746 & \cellcolor{lightred}{-4.48} & 0.580 & \cellcolor{lightred}{-6.30} & 0.435 & \cellcolor{lightred}{-7.64} & 0.319 & \cellcolor{lightred}{-9.10} & 0.313 & \cellcolor{lightred}{-6.85} & 0.568 & \cellcolor{lightred}{-3.90} & 0.809 & \cellcolor{lightred}{-2.06} & 0.772 & \cellcolor{lightred}{-2.16} & 0.725 & \cellcolor{lightred}{-2.68} \\
SD2-ld   & 0.742 & \cellcolor{lightred}{-4.99} & 0.576 & \cellcolor{lightred}{-6.94} & 0.432 & \cellcolor{lightred}{-8.28} & 0.316 & \cellcolor{lightred}{-9.97} & 0.311 & \cellcolor{lightred}{-7.44} & 0.569 & \cellcolor{lightred}{-3.73} & 0.807 & \cellcolor{lightred}{-2.30} & 0.767 & \cellcolor{lightred}{-2.79} & 0.722 & \cellcolor{lightred}{-3.09} \\
SD3-cm   & 0.563 & \cellcolor{lightred}{-27.91} & 0.376 & \cellcolor{lightred}{-39.26} & 0.248 & \cellcolor{lightred}{-47.36} & 0.165 & \cellcolor{lightred}{-52.99} & 0.232 & \cellcolor{lightred}{-30.95} & 0.418 & \cellcolor{lightred}{-29.26} & 0.645 & \cellcolor{lightred}{-21.91} & 0.582 & \cellcolor{lightred}{-26.21} & 0.525 & \cellcolor{lightred}{-29.53} \\
SD3-gc   & 0.715 & \cellcolor{lightred}{-8.46} & 0.553 & \cellcolor{lightred}{-10.67} & 0.409 & \cellcolor{lightred}{-13.17} & 0.297 & \cellcolor{lightred}{-15.37} & 0.305 & \cellcolor{lightred}{-9.23} & 0.552 & \cellcolor{lightred}{-6.60} & 0.791 & \cellcolor{lightred}{-4.24} & 0.746 & \cellcolor{lightred}{-5.44} & 0.695 & \cellcolor{lightred}{-6.71} \\
SD3-ld   & 0.755 & \cellcolor{lightred}{-3.33} & 0.590 & \cellcolor{lightred}{-4.68} & 0.444 & \cellcolor{lightred}{-5.74} & 0.328 & \cellcolor{lightred}{-6.55} & 0.318 & \cellcolor{lightred}{-5.36} & 0.576 & \cellcolor{lightred}{-2.54} & 0.819 & \cellcolor{lightred}{-0.85} & 0.781 & \cellcolor{lightred}{-1.01} & 0.737 & \cellcolor{lightred}{-1.07} \\
\emph{Orig.} & 0.781 & — & 0.619 & — & 0.471 & — & 0.351 & — & 0.336 & — & 0.591 & — & 0.826 & — & 0.789 & — & 0.745 & — \\
\bottomrule
\end{tabular}}
\label{tab:og_with_percent}
\end{table*}

\paragraph{Guidance Scale Analysis}

\begin{figure}[ht]
    \centering
    \includegraphics[width=0.9\columnwidth]{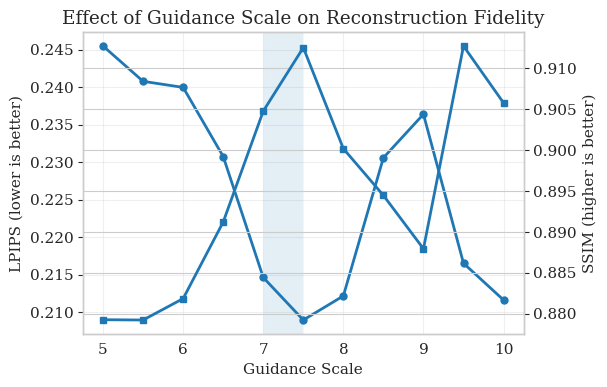}
    \caption{Effect of classifier-free guidance scale on reconstruction fidelity. LPIPS (left axis, lower is better) and SSIM (right axis, higher is better) are shown as a function of guidance over 100 samples. Reconstruction quality improves as guidance increases from 5.0 to approximately 7.0-7.5, after which both metrics exhibit increased variance and non-monotonic behavior, indicating diminishing or unstable gains at higher guidance values.}
    \label{fig:guidance_scale}
\end{figure}

We analyze the effect of classifier-free guidance on reconstruction fidelity using LPIPS and SSIM. As shown in Figure~\ref{fig:guidance_scale}, increasing guidance from low values improves perceptual and structural quality, with LPIPS decreasing and SSIM increasing up to a guidance range of approximately 7.0-7.5. Beyond this range, improvements saturate and become unstable, with no consistent gains at higher guidance scales. This indicates that moderate guidance achieves a stable trade-off between perceptual similarity and structural fidelity, while larger guidance values provide limited benefit.

\subsection{RefCOCOg dataset (Success Case)}
\label{sec:success_cases_refcocog}

\paragraph{Analysis}
RefCOCOg exhibits very strong and consistent reconstruction–caption correlations, comparable to or stronger than those observed on Flickr.
MSE and LPIPS show near-linear negative correlations with caption quality, while PSNR exhibits correspondingly strong positive correlations across BLEU$_\text{avg}$, METEOR, and SBERT.
These trends indicate that reconstruction fidelity within the annotated region is critical for successful referring expression generation.
In contrast, SSIM remains weakly correlated and unstable, suggesting limited sensitivity to semantically relevant degradation.
Across diffusion variants, SD2 and SD3 with Gaussian or low-dimensional degradation perform best, while center-masked reconstructions consistently underperform due to complete removal of the referential target (see Table~\ref{tab:refcocog_recon} and~\ref{tab:refcocog_caption}).

\begin{table}[H]
\centering
\scriptsize
\setlength{\tabcolsep}{5pt}
\renewcommand{\arraystretch}{1.15}
\caption{\textbf{Pixel-level reconstruction metrics across inpainting variants for RefCOCOg.}
MSE $\downarrow$, PSNR $\uparrow$, SSIM $\uparrow$, and LPIPS $\downarrow$ are averaged over the evaluation set.
Lower MSE/LPIPS and higher PSNR/SSIM indicate better reconstruction fidelity.}
\resizebox{\columnwidth}{!}{%
\begin{tabular}{lcccc}
\toprule
\textbf{Variant} & \textbf{MSE} & \textbf{PSNR} & \textbf{SSIM} & \textbf{LPIPS} \\
\midrule
SD1.5-cm & 0.03232 & 16.09 & 0.583 & 0.259 \\
SD1.5-gc & 0.02918 & 15.99 & 0.544 & 0.275 \\
SD1.5-ld & 0.02843 & 16.48 & 0.629 & 0.253 \\
\midrule
SD2-cm & 0.02757 & 17.05 & 0.598 & 0.240 \\
SD2-gc & 0.02529 & 17.11 & 0.590 & 0.244 \\
SD2-ld & 0.02480 & 17.45 & 0.657 & 0.211 \\
\midrule
SD3-cm & 0.02360 & 17.35 & 0.611 & 0.218 \\
SD3-gc & \textbf{0.02082} & 18.38 & 0.582 & 0.187 \\
SD3-ld & 0.02212 & \textbf{18.56} & \textbf{0.614} & \textbf{0.188} \\
\bottomrule
\end{tabular}
}
\label{tab:refcocog_recon}
\end{table}

\begin{table}[H]
\centering
\scriptsize
\setlength{\tabcolsep}{6pt}
\renewcommand{\arraystretch}{1.15}
\caption{\textbf{Captioning performance across inpainting variants for RefCOCOg (QWEN).}
Reported metrics include BLEU average (BLEU-1–4), METEOR, and SBERT cosine similarity.
Higher values indicate better caption quality and semantic alignment.}
\resizebox{\columnwidth}{!}{%
\begin{tabular}{lccc}
\toprule
\textbf{Variant} & \textbf{BLEU$_\text{avg}$} & \textbf{METEOR} & \textbf{SBERT} \\
\midrule
SD1.5-cm & 0.516 & 0.240 & 0.595 \\
SD1.5-gc & 0.572 & 0.279 & 0.580 \\
SD1.5-ld & 0.594 & 0.313 & 0.590 \\
\midrule
SD2-cm & 0.579 & 0.301 & 0.603 \\
SD2-gc & 0.613 & 0.304 & 0.623 \\
SD2-ld & 0.614 & 0.345 & 0.653 \\
\midrule
SD3-cm & 0.642 & 0.337 & 0.657 \\
SD3-gc & 0.638 & 0.347 & \textbf{0.718} \\
SD3-ld & 0.640 & \textbf{0.361} & 0.707 \\
\bottomrule
\end{tabular}
}
\label{tab:refcocog_caption}
\end{table}

\subsection{TRUCE dataset (Success Case)}
\label{sec:success_case_truce}

\paragraph{Analysis.} TRUCE exhibits clear and consistent reconstruction–caption correlations, though lower strength than the two natural-image datasets. MSE and LPIPS maintain negative correlations with caption quality, while PSNR shows positive correlations, with the strongest relationships observed for SBERT similarity (up to $|r| \approx 0.8$). Lexical metrics such as BLEU$_\text{avg}$ and METEOR show more moderate correlations. Across both reconstruction and captioning tasks, SD2-gc and SD3-gc perform best, achieving the strongest captioning scores (Table~\ref{tab:truce_results}) and the lowest reconstruction error (Table~\ref{tab:recon_metrics_truce}) respectively. In contrast, center-mask variants consistently underperform. SSIM shows more stable but moderate correlations compared to observations on natural-image datasets. TRUCE confirms that the reconstruction–caption relationship generalizes beyond natural images, albeit with slightly reduced correlation strength.

\begin{table}[H]
\centering
\scriptsize
\setlength{\tabcolsep}{4pt}
\renewcommand{\arraystretch}{1.15}
\caption{\textbf{Pixel-level reconstruction metrics across inpainting variants for TRUCE.}
MSE $\downarrow$, PSNR $\uparrow$, SSIM $\uparrow$, and LPIPS $\downarrow$ are averaged over all samples.
Lower MSE/LPIPS and higher PSNR/SSIM indicate better reconstruction fidelity.}
\resizebox{\columnwidth}{!}{%
\begin{tabular}{lcccc}
\toprule
\textbf{Variant} & \textbf{MSE} & \textbf{PSNR} & \textbf{SSIM} & \textbf{LPIPS} \\

\midrule
SD1.5-cm & 0.0544 & 15.24 & 0.871 & 0.177 \\
SD1.5-gc & 0.0439 & 15.97 & 0.879 & 0.171 \\

\midrule
SD2-cm & 0.0360 & 17.20 & 0.892 & 0.161 \\
SD2-gc & 0.0235 & 19.67 & 0.925 & 0.128 \\

\midrule
SD3-cm & 0.0158 & 20.10 & 0.945 & 0.096 \\
SD3-gc & \textbf{0.0122} & \textbf{21.87} & \textbf{0.951} & \textbf{0.086} \\
\bottomrule
\end{tabular}
}
\label{tab:recon_metrics_truce}
\end{table}

\begin{table}[H]
\centering
\scriptsize
\setlength{\tabcolsep}{3pt}
\renewcommand{\arraystretch}{1.15}
\caption{\textbf{Captioning performance on the TRUCE dataset (QWEN).}
Reported metrics include BLEU-1–4, METEOR (MET.), ROUGE-L (R-L), and SBERT cosine similarity.
Abbreviations — cm: center mask; gc: Gaussian center.}
\resizebox{\columnwidth}{!}{%
\begin{tabular}{lccccccc}
\toprule
\textbf{Variant} & B1 & B2 & B3 & B4 & MET. & R-L & SBERT \\
\midrule
SD1.5-cm & 0.275 & 0.105 & 0.042 & 0.0260 & 0.195 & 0.190 & 0.47 \\
SD1.5-gc & 0.295 & 0.115 & 0.046 & 0.0280 & 0.210 & 0.205 & 0.48 \\
SD2-cm   & 0.285 & 0.110 & 0.044 & 0.0270 & 0.205 & 0.200 & 0.49 \\
SD2-gc   & \textbf{0.315} & \textbf{0.125} & \textbf{0.049} & \textbf{0.0300} & \textbf{0.225} & \textbf{0.220} & \textbf{0.52} \\
SD3-cm   & 0.305 & 0.120 & 0.047 & 0.0290 & 0.220 & 0.215 & 0.51 \\
SD3-gc   & 0.290 & 0.108 & 0.043 & 0.0265 & 0.208 & 0.202 & 0.50 \\
\bottomrule
\end{tabular}
}
\label{tab:truce_results}
\end{table}

\subsection{ROCOv2 (Qualified Success Case)}
\label{sec:success_case_roco}

\begin{figure}[H]
    \centering
    \includegraphics[width=\columnwidth, trim={680 0 0 0}, clip]{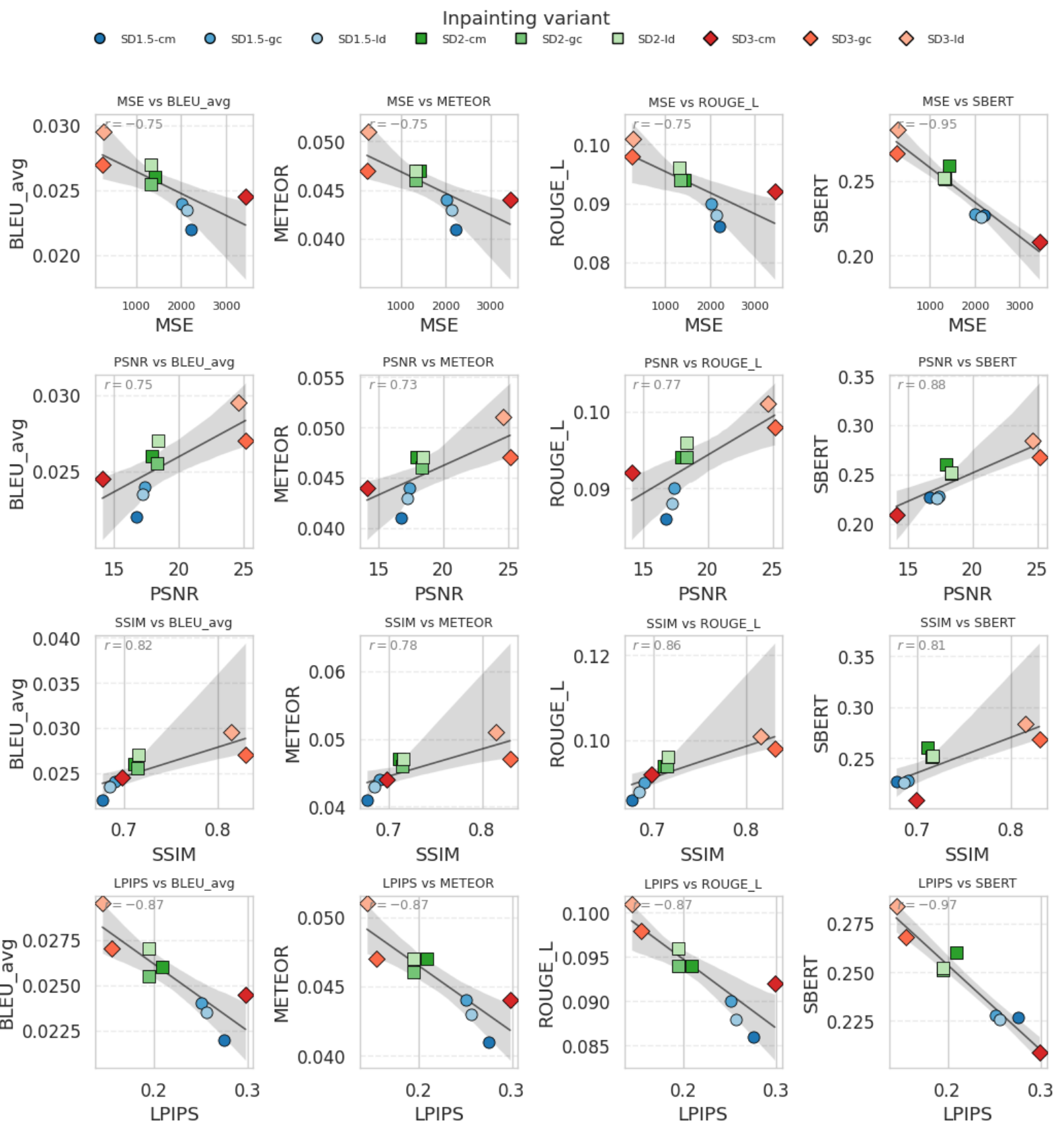}
        \caption{Relationship between reconstruction fidelity and captioning performance across Stable Diffusion variants and BLIP model using the ROCOv2 dataset}
    \label{fig:roco_corr}
\end{figure}

\paragraph{Analysis.}
ROCOv2 exhibits a clear and internally consistent relationship between reconstruction fidelity and caption quality (Figure~\ref{fig:roco_corr}).
Across inpainting variants, MSE and LPIPS show strong negative correlations with all captioning metrics (up to $|r|=0.95$ for SBERT), while PSNR shows correspondingly strong positive correlations, indicating that relative improvements in reconstruction fidelity reliably translate to improved language alignment.
SSIM also demonstrates strong positive correlations across all caption metrics, suggesting that structural preservation is more informative for semantic grounding in medical imagery than in natural-image datasets.
The primary distinction from Flickr and RefCOCOg lies in the absolute scale of reconstruction error, with substantially higher MSE values due to the low contrast and fine-grained structure of medical images rather than a breakdown in semantic alignment.
Despite this scale difference, metric ordering and cross-metric agreement are preserved, and Gaussian-center and low-dimensional masking under SD2 and SD3 consistently achieve the strongest joint reconstruction–caption performance (Table~\ref{tab:reconstruction_metrics_roco_v2} and Table~\ref{tab:rocotab}).

\begin{table}[H]
\centering
\scriptsize
\setlength{\tabcolsep}{4pt}
\renewcommand{\arraystretch}{1.15}
\caption{\textbf{Pixel-level reconstruction metrics across inpainting variants for ROCO V2.}
MSE $\downarrow$, PSNR $\uparrow$, SSIM $\uparrow$, and LPIPS $\downarrow$ are averaged over the evaluation set.
Lower MSE/LPIPS and higher PSNR/SSIM indicate better reconstruction fidelity.}
\resizebox{\columnwidth}{!}{%
\begin{tabular}{lcccc}
\toprule
\textbf{Variant} & \textbf{MSE} & \textbf{PSNR} & \textbf{SSIM} & \textbf{LPIPS} \\

\midrule
SD1.5-cm & 2217.34 & 16.74 & 0.677 & 0.275 \\
SD1.5-gc & \textbf{2012.18} & \textbf{17.41} & \textbf{0.690} & \textbf{0.251} \\
SD1.5-ld & 2132.21 & 17.25 & 0.685 & 0.256 \\

\midrule
SD2-cm & 1417.41 & 17.97 & 0.711 & 0.209 \\
SD2-gc & 1336.09 & 18.34 & 0.715 & 0.195 \\
SD2-ld & \textbf{1320.51} & \textbf{18.39} & \textbf{0.716} & \textbf{0.195} \\

\midrule
SD3-cm & 3442.00 & 14.13 & 0.698 & 0.298 \\
SD3-gc & \textbf{245.75} & \textbf{25.15} & \textbf{0.830} & 0.155 \\
SD3-ld & 276.92 & 24.64 & 0.815 & \textbf{0.145} \\
\bottomrule
\end{tabular}
}
\label{tab:reconstruction_metrics_roco_v2}
\end{table}

\begin{table}[ht]
\centering
\scriptsize
\setlength{\tabcolsep}{3pt}
\renewcommand{\arraystretch}{1.15}
\caption{\textbf{Captioning performance across inpainting variants for the ROCO V2 dataset.}
Reported metrics include BLEU$_\text{avg}$ (average of BLEU-1 and BLEU-2), METEOR (MET.), ROUGE-L (R-L),
supervised SimCSE (sup.), unsupervised SimCSE (unsup.), and SBERT cosine similarity.}
\label{tab:captioning_performance_rocov2}
\resizebox{\columnwidth}{!}{%
\begin{tabular}{lrrrrrr}
\toprule
\textbf{Variant} & \textbf{BLEU$_\text{avg}$} & \textbf{MET.} & \textbf{R-L} & \textbf{sup.} & \textbf{unsup.} & \textbf{SBERT} \\
\midrule
\multicolumn{7}{c}{\textbf{BLIP}} \\
\midrule
SD1.5-cm & 0.022 & 0.041 & 0.086 & 0.322 & 0.250 & 0.227 \\
SD1.5-gc & 0.024 & 0.044 & 0.090 & 0.326 & 0.257 & 0.228 \\
SD1.5-ld & 0.024 & 0.043 & 0.088 & 0.319 & 0.253 & 0.226 \\
SD2-cm   & 0.026 & 0.047 & 0.094 & 0.366 & 0.296 & 0.260 \\
SD2-gc   & 0.026 & 0.046 & 0.094 & 0.365 & 0.288 & 0.251 \\
SD2-ld   & 0.027 & 0.047 & 0.096 & 0.363 & 0.291 & 0.252 \\
SD3-cm   & 0.025 & 0.044 & 0.092 & 0.310 & 0.245 & 0.209 \\
SD3-gc   & 0.027 & 0.047 & 0.098 & 0.397 & 0.329 & 0.268 \\
SD3-ld   & 0.030 & 0.051 & 0.101 & 0.413 & 0.340 & 0.284 \\
% Orig.    & 0.028 & 0.048 & 0.096 & 0.369 & 0.305 & 0.258 \\
\midrule
\multicolumn{7}{c}{\textbf{QWEN}} \\
\midrule
SD1.5-cm & 0.086 & 0.120 & 0.177 & 0.589 & 0.585 & 0.409 \\
SD1.5-gc & 0.086 & 0.118 & 0.175 & 0.590 & 0.588 & 0.404 \\
SD1.5-ld & 0.087 & 0.121 & 0.179 & 0.592 & 0.585 & 0.405 \\
SD2-cm   & 0.092 & 0.125 & 0.187 & 0.613 & 0.608 & 0.433 \\
SD2-gc   & 0.092 & 0.126 & 0.190 & 0.611 & 0.606 & 0.432 \\
SD2-ld   & 0.094 & 0.129 & 0.187 & 0.609 & 0.604 & 0.429 \\
SD3-cm   & 0.094 & 0.124 & 0.186 & 0.597 & 0.595 & 0.409 \\
SD3-gc   & 0.092 & 0.125 & 0.189 & 0.616 & 0.608 & 0.441 \\
SD3-ld   & 0.093 & 0.127 & 0.191 & 0.619 & 0.610 & 0.440 \\
% Orig.    & 0.099 & 0.135 & 0.205 & 0.638 & 0.621 & 0.465 \\
\bottomrule
\end{tabular}}
\label{tab:rocotab}
\end{table}

\section{Failure Cases}
\label{sec:failure_cases}

\begin{figure}[ht]
    \centering
    \includegraphics[width=0.75\columnwidth, trim={300 0 300 30}, clip]{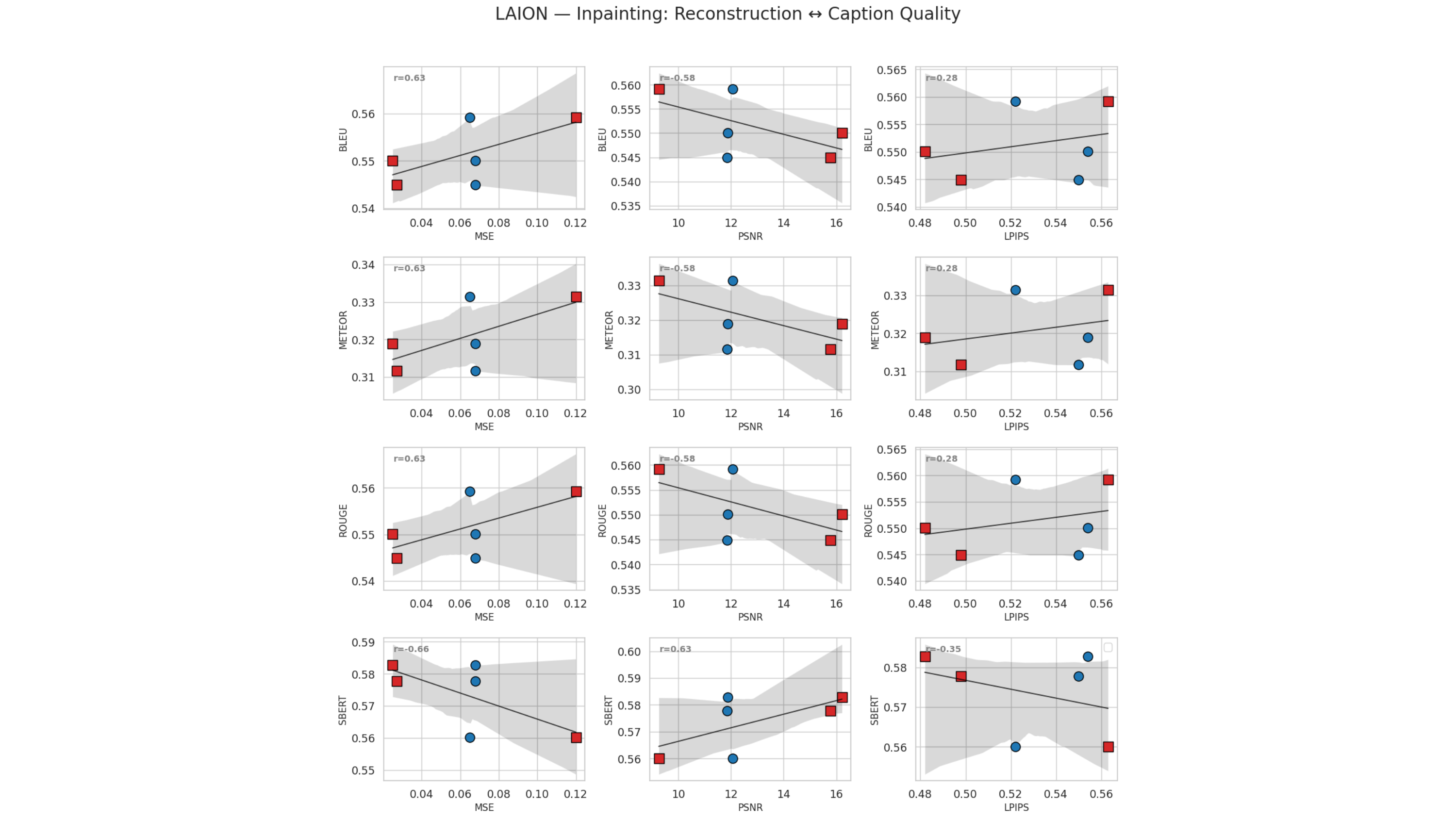}
    \caption{Relationship between reconstruction fidelity and captioning performance across Stable Diffusion variants and LAION model using the GTZAN dataset }
    \label{fig:gtzan_corr}
\end{figure}

\begin{figure}[H]
    \centering
    \includegraphics[width=0.75\columnwidth, trim={22 20 0 130}, clip]{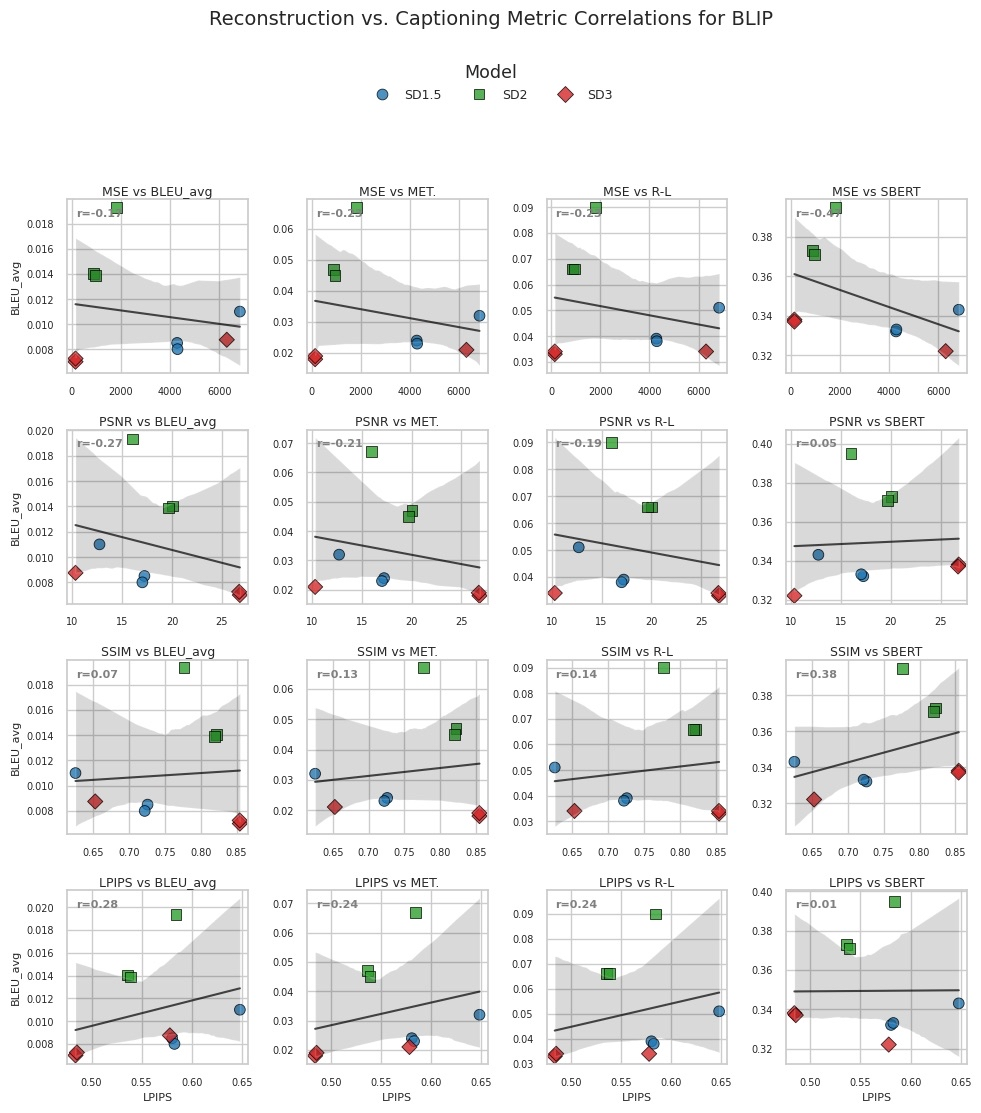}
    \caption{Relationship between reconstruction fidelity and captioning performance across Stable Diffusion variants and BLIP model using the X-ray dataset }
    \label{fig:recon_caption_corr_med}
\end{figure}

Both the GTZAN and X-ray datasets fail to exhibit meaningful reconstruction to caption correlations despite large variations in reconstruction fidelity across inpainting variants (see Figure~\ref{fig:gtzan_corr} and~\ref{fig:recon_caption_corr_med}). This failure is not caused by limited reconstruction diversity. Metrics such as MSE, PSNR, and LPIPS span wide ranges across variants. Instead, the failure arises from caption impoverishment.

We report results using LP-MusicCaps~\cite{doh2023lpmusiccaps} and LAION-CLAP~\cite{elizalde2023clap} and for inpainting we use Riff-ControlNet~\cite{shah2024riffcontrolnet} and  Riffusion~\cite{riffusion2022}. For GTZAN, captions are extremely short, generic, or effectively label-like, often consisting of genre names or minimal descriptors. As a result, captioning models generate nearly identical outputs for clean and heavily corrupted inputs. Caption quality metrics, therefore, remain almost constant, making them insensitive to reconstruction quality. Even large changes in reconstruction fidelity do not translate into measurable differences in caption performance.

A similar pattern is observed for X-ray images, where captions are repetitive and coarse-grained, typically describing global anatomical structures rather than localized visual evidence. The masked regions introduced during inpainting do not consistently overlap with the visual cues emphasized in the captions. Consequently, reconstruction differences have little influence on caption generation, leading to weak or unstable correlations across all metric pairs.

We include these datasets intentionally to test whether reconstruction to caption relationships persist under minimal linguistic variability. Their failure confirms that expressive and semantically rich captions are a necessary condition for reconstruction-aware caption evaluation. When captions lack modality-specific detail, reconstruction fidelity becomes largely irrelevant to downstream language outputs, even under substantial visual perturbations.

% This is an appendix.

\end{document}